\documentclass[10pt,twocolumn,letterpaper]{article}

\usepackage[pagenumbers]{cvpr} 
\usepackage{dsfont}
\usepackage{cuted}
\usepackage{currfile}
\usepackage{caption}

\usepackage{booktabs}
\usepackage{multirow}
\usepackage[table]{xcolor}
\definecolor{offlinegray}{HTML}{E8E8E8}

\newcommand{\best}[1]{\cellcolor{blue!22}\textbf{#1}}
\newcommand{\good}[1]{\cellcolor{blue!12}\underline{#1}}
\newcommand{\fair}[1]{#1}

\definecolor{cvprblue}{rgb}{0.21,0.49,0.74}
\usepackage[pagebackref,breaklinks,colorlinks,allcolors=cvprblue]{hyperref}

\def\confName{CVPR}
\def\confYear{2027}

\title{OCLDA-GS: Object-Centric Local Deformation-Aware Gaussian Streaming}
\author{Ashim Dahal$^{1}$ \quad Nikesh Subedi$^{2}$ \quad Rabab Abdelfattah$^{1}$ \quad Nick Rahimi$^{1}$\\
$^{1}$University of Southern Mississippi, Hattiesburg, MS\\
$^{2}$University of Utah, Salt Lake City, UT\\
{\tt\small \{ashim.dahal, rabab.abdelfattah, nick.rahimi\}@usm.edu, nikesh.subedi@utah.edu}
}

\newcommand{\AD}[1]{\textcolor{black}{#1}}
\newcommand{\ADc}[1]{\textcolor{cyan}{}}

\newcommand{\NSc}[1]{\textcolor{olive}{}}

\begin{document}
\maketitle
\begin{strip}
\centering
\captionsetup{type=figure,hypcap=false}
\begin{minipage}[t]{0.67\textwidth}
    \vspace{0pt}
    \centering
    \includegraphics[width=\linewidth]{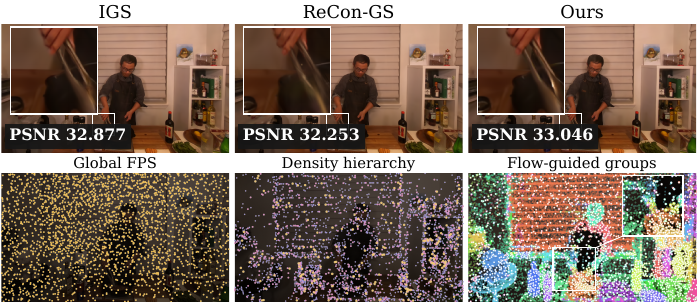}
    \subcaption{Reconstruction and anchor selection.}
    \label{fig:teaser_anchors}
\end{minipage}%
\hfill
\begin{minipage}[t]{0.29\textwidth}
    \vspace{0pt}
    \centering
    \includegraphics[width=\linewidth]{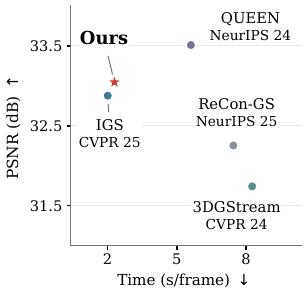}
    \subcaption{Quality--efficiency tradeoff.}
    \label{fig:teaser_pareto}
\end{minipage}
\caption{\textbf{Object-centric anchors for streaming reconstruction.} (a) Selected N3DV renders with dataset-average PSNR (dB). Below, anchors selected by respective methods are shown, colors show object groups. (b) Quality versus reconstruction time (in seconds per frame) in the N3DV dataset.}
\label{fig:teaser}
\captionsetup{hypcap=true}
\end{strip}

\begin{abstract}
Streaming 3D Gaussian Splatting (3DGS) allows for online reconstruction of dynamic scenes from multi-view videos for Free Viewpoint Video (FVV). \NSc{A sentence on why this is challenging;/why is this important; application} Current time and storage efficient methods rely on selecting a subset of anchor Gaussians and neighborhood interpolation for fast streamable 3DGS. However, these methods can overlook object-specific motion and introduce inconsistencies near object boundaries. We propose OCLDA-GS, an object-centric streaming framework that dynamically reallocates object and motion-aware Gaussian anchors \AD{to allocate anchor Gaussian support for objects undergoing large motion while at the same time, restricting the gaussian neighborhood within object groups for more precise object boundaries}\NSc{to do ...; to allocate more anchors to high motion areas/object which are likely to be changing}. Furthermore, it estimates object-wise rigid transformations (e.g. a hand moving) \NSc{to approximate how an object might move; hand moving to the right moves the fingers along with it} \ADc{I think this one doesn't need a to as its more so contributing to the third piece next} and \AD{leverages these estimates to learn per-anchor} \AD{ local deformation correction for small deformations (e.g. wrinkles on a shirt)}\NSc{to model local changes; clothes on the arm/hand might have wrinkles or local deformation}. Then, we conduct experiments on three dynamic multi-view video datasets and demonstrate improvements over the previous baselines alongside improved quality between refinement timesteps. Under matched refinement budgets, our method improves PSNR over Instant Gaussian Stream (IGS) by 0.169 dB on N3DV, 0.202 dB on Meeting Room \AD{and +0.199 on Google Immersive}, while reducing the mean PSNR\xspace \AD{drop}\NSc{remove double negative} between refinement timesteps by 9.2\% and 13.8\% in N3DV and Meeting Room, respectively.

\noindent\emph{Code is available via the supplementary zip.}
\end{abstract}    
\section{Introduction}
3D Gaussian Splatting (3DGS)~\cite{kerbl2023gaussians} allows for photorealistic reconstruction from RGB images and camera poses using 3D Gaussians as primitives. \NSc{applications/more methods have done this and that enabling this and that application}
4D Gaussian Splatting (4DGS) or Dynamic Gaussian Splatting \AD{adds spatiotemporal deformation in 3DGS}\NSc{define; generalizes/ makes 3DGS time variant...; then application}. This enables realtime Free Viewpoint Video (FVV) for novel view synthesis~\cite{wu2024gaussians,yang2024realtime4dgs,luiten2024dynamic}. FVV has applications in VR, immersive communication and sports viewing~\cite{broxton2020immersive,yan2025instantgaussianstreamfast}. Current Dynamic Gaussian Splatting methods can be divided into two broad categories: offline and online~\cite{wu2024gaussians,sun20243dgstream,girish2024queen}. Offline methods optimize spatiotemporal deformation by learning the motion features throughout the entire video and emphasizes quality and refinement taking significant time and compute in the process~\cite{yang2024deformable3dgs,yang2024realtime4dgs}. In the online streaming setting, the scene representation is updated as each set of new multi-view frames arrive and the central concern is to make that update computationally cheap while maintaining image fidelity~\cite{sun20243dgstream,girish2024queen}. 

Recent time and storage-efficient Gaussian streaming methods rely on selection of subset of Gaussians called anchors, whose motion features are then interpolated to update the scene representation. Methods like IGS~\cite{yan2025instantgaussianstreamfast} and ReconGS \cite{fu2025recongs} work under fixed computational budget controlled by the number of anchors and refinement per step or keyframes making them extremely effective for real-time dynamic scene streaming under limited computation bandwidth. IGS performs farthest point sampling (FPS) and periodic refinement on pre-specified timesteps called keyframes, while ReConGS~\cite{fu2025recongs} organizes anchors into an adaptive hierarchy for compact motion optimization and performs per-step refinements. 

\NSc{These methods assume 3d deformation/motion share similar motion characteristic to its neighborhood.}\AD{These methods assume neighboring Gaussians share the 3D deformation trajectories of each other.}
\ADc{Sparse deformation via anchors propagates transformation by interpolating spatially weighted neighborhoods}~\cite{sumner2007embedded,huang2024scgs}.\ADc{ However, geometric proximity does not necessitate shared motion. }\NSc{However, this is not true in object boundaries.} \AD{ However, this assumption fails for Gaussians near object boundaries.} For example, in the N3DV dataset (~\cref{fig:teaser_anchors}), when the chef's hand moves near a stationary table, deformation via interpolated anchor neighborhood can induce incompatible motion in Gaussians belonging to the stationary table. 

\NSc{ Maybe remove: Geometry-based anchor coverage also do not allocate sufficient anchors to regions undergoing substantial motion.} \AD{Existing anchor selection methods like FPS or hierarchical sampling also may not reflect the motion distribution of the scene accurately. Under a limited anchor budget, these methods can allocate excessive support to stationary regions.} Some related object and motion-aware control-point methods address parts of this challenge~\cite{he2025h3ddgs,chen2025comgs}, but maintaining accurate, object-consistent motion with limited refinement remains challenging. Furthermore, motion within an object often comprises
a shared rigid component and local non-rigid deformations
(e.g., a moving body with wrinkling clothes).
Accounting for both components is important for accurate
motion prediction under limited refinement budgets.
\ADc{Add missing motivation on the two claims from abstract: objectwise transformation + per-anchor residual}

We introduce Object-Centric Local Deformation-Aware Gaussian Streaming \emph{(OCLDA-GS)}\NSc{change name}. OCLDA-GS jointly addresses anchor selection, object-aware motion propagation, and local deformation correction. First, OCLDA-GS uses optical flow~\cite{teed2020raft} \NSc{cite older?} observed over each completed keyframe interval to redistribute a fixed anchor budget across object groups, increasing support for groups with larger observed motion. Second, it restricts motion propagation to anchors within the same group, preventing cross-group interpolation for supported Gaussians. Finally, it fits a rigid transformation to the predicted anchor correspondences of each group and predicts an additional local deformation correction.

More broadly, OCLDA-GS highlights object-aware motion modeling
as a promising direction for improving the quality--efficiency
trade-off in Gaussian streaming. Extensive experiments on \AD{three} \NSc{three; change in abstract too} datasets demonstrate competitive
reconstruction quality and streaming efficiency in both in-domain
and cross-domain settings. Temporal analyses against online streaming method like Instant Gaussian Stream (IGS) further show reduced
quality degradation \NSc{double negation remove?} between keyframes.
In summary, our contributions are:
\begin{itemize}
    \item \AD{\textbf{Flow-guided anchor selection:}
    We dynamically redistribute a fixed anchor budget
    across object groups based on observed optical flow,
    providing greater motion support for objects
    undergoing larger motion.}

    \item \AD{\textbf{Object-constrained motion propagation:}
    We restrict anchor-based motion interpolation to
    within object groups, preventing incompatible motion
    from propagating across object boundaries.}

    \item \AD{\textbf{Rigid-conditioned deformation correction:}
    We leverage object-wise rigid motion estimates to
    learn per-anchor local deformation corrections, accounting
    for local non-rigid deformations within objects
    without additional refinement.}
\end{itemize}
\section{Related Work}
\label{sec:related_work}

\noindent\textbf{Dynamic novel view synthesis.}
NeRF established neural radiance fields as a representation for novel
view synthesis~\cite{mildenhall2020nerf}. Dynamic extensions model
scene changes through deformation fields, scene flow, and
higher-dimensional embeddings
~\cite{pumarola2021dnerf,park2021nerfies,li2021nsff,park2021hypernerf}.
HexPlane and K-Planes improve efficiency through factorized
spatiotemporal features
~\cite{cao2023hexplane,fridovichkeil2023kplanes}.
With an explicit Gaussian representation and differentiable
rasterization, 3DGS enables high-quality real-time
rendering~\cite{kerbl2023gaussians}.
Dynamic Gaussian methods extend this representation using learned
deformation fields and spatiotemporal Gaussian formulations
~\cite{yang2024deformable3dgs,wu2024gaussians,yang2024realtime4dgs,li2024spacetime}.
These advances improve dynamic scene representation and rendering,
while online reconstruction additionally requires efficient updates
as new observations arrive.

\noindent\textbf{Efficient Streaming.}
StreamRF and ReRF investigate incremental learning and residual
representations for dynamic radiance fields
~\cite{li2022streaming,wang2023rerf}, while NeRFPlayer and VideoRF
explore decomposed fields and feature streams for compact video
representation and playback
~\cite{song2023nerfplayer,wang2024videorf}.
For Gaussian-based online reconstruction, 3DGStream trains scene
updates on the fly~\cite{sun20243dgstream}, and QUEEN uses sparse,
quantized Gaussian attribute residuals~\cite{girish2024queen}.
Complementary compression approaches, including 4DGC and GIFStream,
target compact representations and efficient video delivery
~\cite{hu20254dgc,li2025gifstream}.
HiCoM and ReCon-GS use hierarchical motion representations to
reduce update and storage costs~\cite{gao2024hicom,fu2025recongs}.
IGS instead employs a generalizable anchor-driven motion predictor
with periodic Gaussian refinement at keyframes
~\cite{yan2025instantgaussianstreamfast}.
OCLDA-GS builds on this inference-based setting to improve motion
support and Gaussian deformation.

\begin{figure*}
    \centering
    \includegraphics[width=0.95\linewidth]{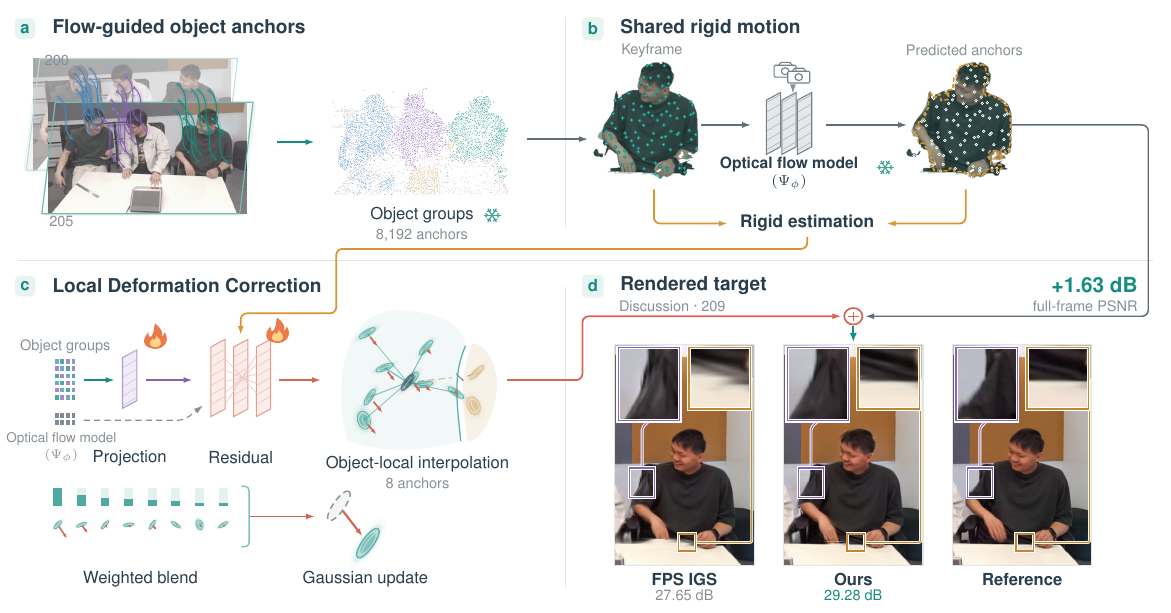}
\caption{\textbf{Overview of our method.}
(a) Flow-guided anchor selection within object groups.
(b) Object-wise rigid motion estimation from AGM-Net predictions.
(c) Learned local deformation corrections propagated to Gaussians
within each object.
(d) Improved rendering quality compared to \AD{Farthest Point Sampling (FPS)}\NSc{full form}-based IGS.}
\label{fig:method}
\end{figure*}

\noindent\textbf{Structured motion and object-aware deformation.}
Embedded deformation represents motion through blended local
transformations~\cite{sumner2007embedded}, and SC-GS uses sparse
control points with learned interpolation and local rigidity
regularization~\cite{huang2024scgs}.
Dynamic 3D Gaussians encourage coherent tracking through local
motion constraints~\cite{luiten2024dynamic}.
MotionGS incorporates optical-flow-based motion guidance, while
MoDec-GS uses global-to-local motion decomposition
~\cite{zhu2024motiongs,kwak2025modec}.
Closer to object-aware streaming, H3D-DGS combines object-wise
motion processing with flow-assisted control
points~\cite{he2025h3ddgs}, and ComGS uses motion-sensitive
keypoints and a motion-aware influence field~\cite{chen2025comgs}.
Building on these ideas, OCLDA-GS combines motion-aware anchor
allocation, explicit group-restricted propagation, and learned
rigid-motion-conditioned correction within keyframe-based streaming.

\section{Method}
\AD{We introduce Object-Centric Local Deformation-Aware Gaussian Streaming (OCLDA-GS). We present a visual overview of our method in \cref{fig:method} and describe it component by component in the following subsections.}
\paragraph{Preliminaries.}
\label{sec:prelim}
We follow the paradigm of keyframe to target frame prediction for streaming \cite{yan2025instantgaussianstreamfast}. This paradigm propagates Gaussian motion from the latest refined
keyframe $k$ to a target frame $t$ using a sparse set
of \AD{Gaussian} anchors. Predicted Gaussian
representation is refined periodically at keyframes, thereby removing computation for per-frame refinement.
Let $\mathcal{G}^{k}=\{G_i^{k}\}_{i=1}^{N}$
denote the Gaussian representation at keyframe $k$,
where each Gaussian is parameterized by its 3D center
$\boldsymbol{\mu}_i^{k}\in\mathbb{R}^{3}$,
\NSc{add optical flow notation here}
orientation quaternion $\mathbf{q}_i^{k}\in\mathbb{R}^{4}$,
and other geometric and appearance attributes.

Let $M \in \mathcal{G}^k$ anchor Gaussians be indexed by
$\mathcal{J}=\{1,\ldots,M\}$.
Let $\rho(j)$ map anchor $j$ to its corresponding
Gaussian index, such that its source position is
$\mathbf{a}_j^{k}=\boldsymbol{\mu}_{\rho(j)}^{k}$.
Given the target frame $t$, a pretrained optical flow conditioned model \NSc{notation for model} \AD{$(\Psi_\phi(.))$} predicts motion features for the selected anchors,
which are interpolated to obtain per-Gaussian features
$\mathbf{f}_i^{k\rightarrow t}$.
A linear motion head $h$ decodes the translation
and rotation updates:
\begin{equation}
\begin{gathered}
(\Delta\boldsymbol{\mu}_i^{k\rightarrow t},
\Delta\mathbf{q}_i^{k\rightarrow t})
= h(\mathbf{f}_i^{k\rightarrow t}),\\
\widehat{\boldsymbol{\mu}}_i^{t}
= \boldsymbol{\mu}_i^{k}
+ \Delta\boldsymbol{\mu}_i^{k\rightarrow t},\\
\widehat{\mathbf{q}}_i^{t}
= \operatorname{norm}
(\mathbf{q}_i^{k}
\otimes\Delta\mathbf{q}_i^{k\rightarrow t}).
\end{gathered}
\label{eq:igs_motion}
\end{equation}
Here, $\otimes$ denotes quaternion multiplication
and $\operatorname{norm}$ denotes quaternion normalization.
The predicted target position of anchor $j$ is
$\widehat{\mathbf{a}}_j^{t}
=\widehat{\boldsymbol{\mu}}_{\rho(j)}^{t}$.

\paragraph{Object Groups.}
We first obtain object-level
segmentation of the Gaussian representation.
\NSc{Introduce Object Semantic feature here; used for segmantation here and local deformation later} \AD{Let $\mathbf{s}_i$ denote the object semantic
embedding feature of Gaussian $i$, and let $\mathbf{c}_o$
denote the mean semantic embedding of anchors
in $\mathcal{A}_o$.}
 Each Gaussian $G_i^k$ is assigned an object-group
label $g_i\in\{-1,0,\ldots,O-1\}$, where $O$
denotes the number of identified objects and
$-1$ indicates an unknown group.
Each anchor $\mathbf{a}_j^k$ has  object-group
label $g_{\rho(j)}$.
The anchor indices belonging to object $o$ are
\begin{equation}
\mathcal{A}_o =
\{j\in\mathcal{J}\mid g_{\rho(j)}=o\}.
\label{eq:object_anchors}
\end{equation}


\paragraph{Anchor Reselection.}
\NSc{Add motivation} \AD{We address the problem of anchor allocation on static regions by periodic anchor redistribution during keyframe refinement.} 
We redistribute the fixed anchor budget $M$ across object groups based on
their size and observed motion.
For each group $o$, we estimate a normalized motion
score $\omega_o^k\in[0,1]$ from optical flow over
the preceding refinement interval, compensating
for global image motion and filtering unreliable
correspondences. Specifically, let $d_{o,v}^k$ be the 75th percentile
of flow magnitudes within group $o$ in camera view $v$,
after subtracting the component-wise median flow
over reliable image pixels. We compute
\begin{equation}
\begin{gathered}
r_o^k =
\operatorname*{median}_{v\in\mathcal V_o^k}
\frac{[d_{o,v}^k-\tau]_+}{D_v},
\\
\omega_o^k =
\min\!\left(1,\frac{r_o^k}
{Q_{0.95}(\{r_g^k:r_g^k>0\})}\right).
\end{gathered}
\label{eq:motion_score}
\end{equation}
where $D_v$ is the image diagonal,
$\tau=0.5$ pixels, $[x]_+=\max(x,0)$, and
$Q_{0.95}$ denotes the 95th percentile.
The set $\mathcal V_o^k$ includes input views with
at least eight distinct reliable, visible projected
pixels for the group.
Let $N_o^k$ denote the number of Gaussians in group $o$.
We assign each group an allocation weight
\begin{equation}
\beta_o^k = \sqrt{N_o^k}(1+\omega_o^k).
\label{eq:flow_allocation}
\end{equation}
The anchor budget is distributed proportionally
to these weights, subject to group population limits.
Within each group, anchors are selected using 3D
farthest-point sampling on the refined Gaussian
centers.

\paragraph{\AD{Object-level} Rigid Motion Estimation.}
For each object $o$, we estimate a rigid transformation
that best approximates its motion predicted by the frozen optical flow model.
Let $\mathcal{Q}_k$ denote the Gaussians which participate in deformation at keyframe $k$. We consider the subset of
object anchors
$\mathcal{F}_o =
\{j\in\mathcal{A}_o \mid \rho(j)\in\mathcal{Q}_k\}$.
Using the source and predicted anchor positions,
we estimate an object-level rotation
$\mathbf{R}_o\in SO(3)$ and translation
$\mathbf{t}_o\in\mathbb{R}^3$ by solving
\begin{equation}
\begin{aligned}
(\mathbf{R}_o,\mathbf{t}_o)
&= \underset{\substack{
\mathbf{R}\in SO(3), 
\mathbf{t}\in\mathbb{R}^3
}}{\arg\min} \sum_{j\in\mathcal{F}_o}
\left\|
\mathbf{R}\mathbf{a}_j^k+\mathbf{t}
-\widehat{\mathbf{a}}_j^t
\right\|_2^2.
\end{aligned}
\label{eq:rigid_fit}
\end{equation}
This least-squares problem is solved analytically
using the Kabsch algorithm \cite{kabsch1976solution}.
The resulting transformation approximates the dominant
rigid motion of each object.

\paragraph{\AD{Local Deformation.} \NSc{local deformation}}
Given the estimated rigid transformation
$(\mathbf{R}_o,\mathbf{t}_o)$, we compute the
difference between the pretrained optical flow predicted anchor positions
at frame $t$ and those obtained by applying the
fitted rigid transformation to their positions
at keyframe $k$.
For each anchor $j\in\mathcal{A}_o$,
\begin{equation}
\mathbf{d}_j^{k\rightarrow t}
=
\mathbf{R}_o^\top
\left(
\widehat{\mathbf{a}}_j^t
-
(\mathbf{R}_o\mathbf{a}_j^k+\mathbf{t}_o)
\right).
\label{eq:rigid_residual}
\end{equation}
The inverse rotation expresses this displacement
in the source-frame axes.

We define the centered anchor position as
$\mathbf{p}_j=\mathbf{a}_j^k-\bar{\mathbf{a}}_o^k$. \NSc{replace SAGA with previous notation}
We condition a shared residual network on the
frozen \AD{optical flow model ($\Psi_\phi$)} features, \NSc{same here; these detail go to implementation detail} anchor motion feature, encoded
anchor position, rigid-relative displacement\NSc{better name},
semantic features, and temporal offset:
\begin{equation}
\begin{gathered}
\mathbf{z}_j^{k\rightarrow t} =
\big(
\mathbf{f}_j^{k\rightarrow t},
\phi(\mathbf{p}_j),
\mathbf{d}_j^{k\rightarrow t},
\mathbf{s}_{\rho(j)}, \mathbf{c}_o,
\tau^{k\rightarrow t}
\big).
\end{gathered}
\label{eq:residual_features}
\end{equation}
where $\phi$ denotes positional encoding and
$\tau^{k\rightarrow t}$ is the normalized temporal
offset.

The residual network $D_\theta$ predicts a
anchor-level translation correction:
\ADc{Ablation without tanh and epsilon}
\begin{equation}
\boldsymbol{\delta}_j^{k\rightarrow t}
=
\begin{cases}
D_\theta(\mathbf{z}_j^{k\rightarrow t})
, & j\in\mathcal{F}_o,\\
\mathbf{0}, & \text{otherwise}.
\end{cases}
\label{eq:anchor_correction}
\end{equation}



\paragraph{Motion and Correction Propagation.}
To prevent cross-object motion mixing, we restrict
the anchors contributing to $\mathbf{f}_i^{k\rightarrow t}$
in \cref{eq:igs_motion} to those belonging to
the same object group as Gaussian $G_i^k$.
For each Gaussian with a known group label $g_i$,
we define its neighborhood as the eight nearest
same-group anchors:
\begin{equation}
\mathcal{N}_8(i) =
\operatorname{KNN}_8
\left(
\boldsymbol{\mu}_i^k,
\{\mathbf{a}_j^k \mid j\in\mathcal{A}_{g_i}\}
\right).
\label{eq:group_neighbors}
\end{equation}
We assign larger weights to anchors closer to the Gaussian
and normalize them within $\mathcal{N}_8(i)$:
\begin{equation}
w_{ij}
=
\frac{\exp\!\left(-\|\boldsymbol{\mu}_i^k-\mathbf{a}_j^k\|_2\right)}
{\sum_{l\in\mathcal{N}_8(i)}
\exp\!\left(-\|\boldsymbol{\mu}_i^k-\mathbf{a}_l^k\|_2\right)}.
\label{eq:interpolation_weights}
\end{equation}
These weights interpolate the anchor motion features
to obtain $\mathbf{f}_i^{k\rightarrow t}$ in
\cref{eq:igs_motion}, and propagate the learned corrections:
\begin{equation}
\boldsymbol{\delta}_i^{k\rightarrow t,\mathrm{loc}}
=
\sum_{j\in\mathcal{N}_8(i)}
w_{ij}
\boldsymbol{\delta}_j^{k\rightarrow t}.
\label{eq:correction_interpolation}
\end{equation}
For supported Gaussians $i\in\mathcal{Q}_k$, we rotate
the correction into world coordinates and add it to
the position predicted by $\Psi_\phi$:
\begin{equation}
\boldsymbol{\mu}_i^{t,\mathrm{pred}}
=
\widehat{\boldsymbol{\mu}}_i^t
+
\mathbf{R}_{g_i}
\boldsymbol{\delta}_i^{k\rightarrow t,\mathrm{loc}}.
\label{eq:final_correction}
\end{equation}

\begin{table*}[!t]
\centering
\small
\setlength{\tabcolsep}{1pt}
\renewcommand{\arraystretch}{1.2}
\caption{Quantitative comparison on N3DV, Meeting Room, and Google Immersive. Average is the equal-weight mean of the three displayed dataset results. Gray rows are offline methods; the remaining rows are online methods. Darker blue indicates the best, and second-best online results. Offline quality includes published reference results.}
\label{tab:main_results}
\resizebox{\linewidth}{!}{%
\begin{tabular}{lcccccccccccccc}
\toprule
\multirow{2}{*}{Method} & \multirow{2}{*}{\shortstack{\# Params}} & \multirow{2}{*}{\shortstack{\# Online\\refinements}} & \multicolumn{3}{c}{N3DV} & \multicolumn{3}{c}{Meeting Room} & \multicolumn{3}{c}{Google Immersive} & \multicolumn{3}{c}{Average} \\
\cmidrule(lr){4-6}
\cmidrule(lr){7-9}
\cmidrule(lr){10-12}
\cmidrule(lr){13-15}
 & & & PSNR $\uparrow$ & DSSIM $\downarrow$ & s/frm $\downarrow$ & PSNR $\uparrow$ & DSSIM $\downarrow$ & s/frm $\downarrow$ & PSNR $\uparrow$ & DSSIM $\downarrow$ & s/frm $\downarrow$ & PSNR $\uparrow$ & DSSIM $\downarrow$ & s/frm $\downarrow$ \\
\midrule
\multicolumn{15}{l}{\textit{Offline methods}} \\
\rowcolor{offlinegray} RT 4DGS~\cite{yang2024realtime4dgs} & 0 &  & 32.010 & 0.014 &  & 30.470 & 0.025 &  & 26.760 & 0.030 &  & 29.747 & 0.023 &  \\
\rowcolor{offlinegray} 4D-GS~\cite{wu2024gaussians} & 2.46M &  & 31.150 & 0.016 &  & 31.275 & 0.024 &  & 23.990 & 0.079 &  & 28.805 & 0.040 &  \\
\rowcolor{offlinegray} STG~\cite{li2024spacetime} & 90 &  & 32.050 & 0.026 &  & 29.510 & 0.034 &  & 29.200 & 0.042 &  & 30.253 & 0.034 &  \\
\rowcolor{offlinegray} Swift4D~\cite{wu2025swift4d} & 31.92M &  & 32.230 & 0.014 &  & 31.306 & 0.022 &  & 23.482 & 0.060 &  & 29.006 & 0.032 &  \\
\midrule
\multicolumn{15}{l}{\textit{Online methods}} \\
ReConGS~\cite{fu2025recongs} & 0 & 2,994 & 32.253 & 0.021 & 7.45 & \good{30.425} & 0.028 & 4.44 & 21.890 & 0.103 & {7.95} & 28.189 & 0.051 & {6.61} \\
3DGStream~\cite{sun20243dgstream} & 1.99M & 2,994 & 31.740 & 0.022 & 8.27 & \best{30.686} & 0.027 & 5.54 & 22.381 & \good{0.079} & 12.62 & 28.269 & {0.043} & 8.81 \\
QUEEN~\cite{girish2024queen} & 345.01M & 2,994 & \best{33.512} & \best{0.018} & {5.61} & 29.126 & 0.027 & {2.25} & \best{25.820} & \best{0.046} & 15.23 & \best{29.486} & \best{0.030} & 7.70 \\
IGS~\cite{yan2025instantgaussianstreamfast} & 21.90M & 594 & {32.877} & {0.020} & \best{2.00} & 30.088 & 0.028 & \best{1.96} & {23.599} & {0.081} & \best{4.55} & {28.855} & {0.043} & \best{2.84} \\
Ours & 27.27M & 594 & \good{33.046} & \good{0.019} & \good{2.30} & 30.290 & 0.027 & \good{2.18} & \good{23.798} & 0.081 & \good{5.50} & \good{29.045} & \good{0.042} & \good{3.33} \\
\bottomrule
\end{tabular}}
\end{table*}

\paragraph{Training Objective.}
We train the residual network using a photometric
reconstruction loss with magnitude and smoothness
regularization:
\begin{equation}
\mathcal{L} =
\mathcal{L}_{\mathrm{photo}}
+ \lambda_{\mathrm{mag}}\mathcal{L}_{\mathrm{mag}}
+ \lambda_{\mathrm{smooth}}\mathcal{L}_{\mathrm{smooth}}.
\label{eq:total_loss}
\end{equation}
Let $\widehat{I}$ denote the rendered image
of the corrected Gaussian representation and
$I$ the ground-truth image. The photometric loss is
\begin{equation}
\begin{aligned}
\mathcal{L}_{\mathrm{photo}}
=&\ \alpha \|\widehat{I}-I\|_1+(1-\alpha)(1-\operatorname{SSIM}(\widehat{I},I)).
\end{aligned}
\label{eq:photometric_loss}
\end{equation}
For the eligible anchor set
$\mathcal{F}=\bigcup_o\mathcal{F}_o$,
let $\mathbf{u}_j =
\boldsymbol{\delta}_j^{k\rightarrow t}/\varepsilon$.
The regularization terms are
\begin{equation}
\begin{aligned}
\mathcal{L}_{\mathrm{mag}}
&= \frac{1}{|\mathcal{F}|}
\sum_{j\in\mathcal{F}}\|\mathbf{u}_j\|_2^2,\\
\mathcal{L}_{\mathrm{smooth}}
&= \frac{1}{Z}
\sum_{(j,l)\in\mathcal{E}}
\gamma_{jl}\|\mathbf{u}_j-\mathbf{u}_l\|_2^2,
\end{aligned}
\label{eq:regularization}
\end{equation}
where $\mathcal{E}$ connects neighboring
same-object anchors, $\gamma_{jl}$ denotes
their spatial-semantic affinity, and
$Z=\max(\sum_{(j,l)\in\mathcal{E}}\gamma_{jl},\eta)$
is the normalization factor.

\section{Experiments}
\subsection{Configuration and Implementation Details}
\paragraph{Datasets.}
We evaluate on three datasets: N3DV~\cite{li2022neural3dvideo},
Meeting Room~\cite{li2022streaming}, and
Google Immersive~\cite{broxton2020immersive}.
We train the residual heads on four N3DV scenes
(Coffee Martini, Cook Spinach, Flame Steak, and Flame Salmon)
and two Google Immersive scenes (Dog and Truck).
Cut Roasted Beef and Sear Steak from N3DV, and Welder and
Alexa Meade Face Paint~1 from Google Immersive, are held out
for testing. Cross-dataset evaluation uses Discussion and
Trimming from Meeting Room. 

\vspace{-0.4cm}

\paragraph{Baselines.}
We compare against four online methods:
IGS~\cite{yan2025instantgaussianstreamfast},
ReConGS~\cite{fu2025recongs},
3DGStream~\cite{sun20243dgstream}, and
QUEEN~\cite{girish2024queen}.
We reproduce these methods using their released implementations
and evaluate them on common held-out scenes, cameras, and frames.
For context, we also include four offline methods:
Real-Time 4DGS~\cite{yang2024realtime4dgs},
4D-GS~\cite{wu2024gaussians},
STG~\cite{li2024spacetime}, and
Swift4D~\cite{wu2025swift4d}. Offline rows are shaded gray and
serve as contextual references. 

\begin{figure*}[!t]
    \centering
    \includegraphics[width=\linewidth]{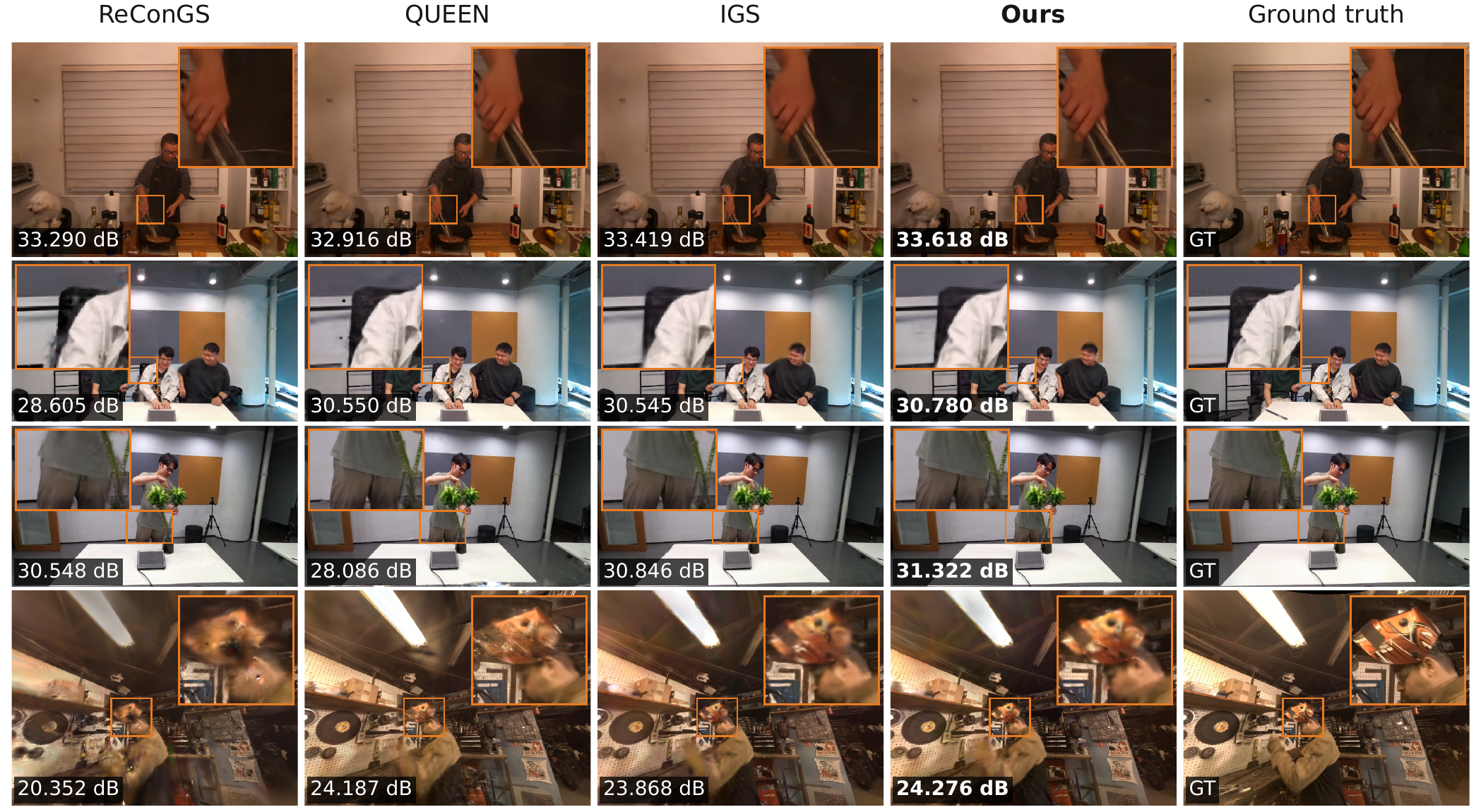}
    \caption{\textbf{Qualitative comparison of online methods across three datasets.}
    Each row uses the same frame, viewpoint, and crop for all methods.
    Orange boxes mark regions enlarged in the insets, highlighting
    motion alignment and object boundaries.}
    \label{fig:qualitative}
\end{figure*}

\begin{figure*}[t]
    \centering
    \includegraphics[width=\linewidth]{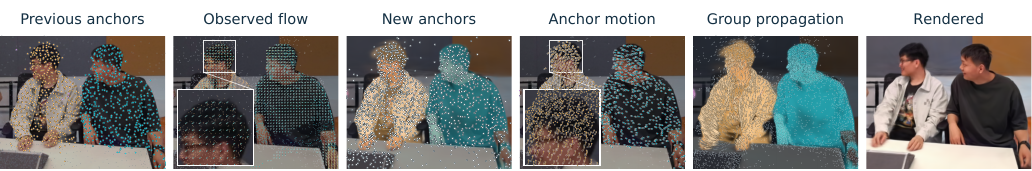}
\caption{\textbf{Flow-guided anchor selection and temporal propagation.}
Flow guides anchor reselection within a fixed budget.
Colors denote SAGA groups; white points mark selected anchors.
Markers and trails visualize anchor and Gaussian motion.
Ours shows the refined target keyframe from an input-camera view.
Reselection and refinement occur only at keyframes.}
    \label{fig:motion_propagation}
\end{figure*}

\paragraph{Implementation Details.}
Our implementation uses PyTorch and runs on an NVIDIA A100
GPU with 80\,GB memory. $\Psi_\phi$ is the frozen AGM-Net from IGS~\cite{yan2025instantgaussianstreamfast}. It receives four input views,
and we use $M=8192$ motion anchors with eight same-group
neighbors per Gaussian, as defined in
\cref{eq:group_neighbors}.
The residual network $D_\theta$ contains two hidden layers
of width 256 with SiLU activations. We use SAGA~\cite{cen2025saga} to segment the Gaussian splat at timestep 0. Details on SAGA grouping are provided in the supplementary.
We optimize \cref{eq:total_loss} using AdamW with a learning
rate of $4\times10^{-4}$ and weight decay of $0.05$,
setting $\alpha=0.8$, $\lambda_{\mathrm{mag}}=0.01$,
and $\lambda_{\mathrm{smooth}}=0.001$.
During streaming, Gaussian refinement is performed every
five frames, using 100 updates for N3DV, 200 for Meeting Room,
and 300 for Google Immersive.
IGS and Ours share the same initialization and refinement
budget for each scene.

\vspace{-0.5cm}
\paragraph{Evaluation Metrics.}
We report PSNR (dB)~\cite{wang2009mse} and structural
dissimilarity, $\mathrm{DSSIM}=(1-\mathrm{SSIM})/2$
\cite{wang2004ssim}, to measure reconstruction quality.
Seconds per frame (s/frm) denotes the average online
reconstruction time per target frame, including periodic
keyframe refinement and excluding initial Gaussian
reconstruction. Higher PSNR and lower DSSIM and s/frm
indicate better performance.

\subsection{Results}

\paragraph{In-domain Evaluation.}
We evaluate on held-out scenes from N3DV and Google Immersive for the in-domain evaluation.
As shown in \cref{tab:main_results}, our method achieves
33.046\,dB on N3DV and 23.798\,dB on Google Immersive,
improving over IGS by 0.169\,dB and 0.199\,dB, respectively.
Our method obtains the second-highest PSNR among the online
methods on both datasets, outperforming IGS, ReConGS, and
3DGStream, while QUEEN achieves the highest PSNR.

\begin{table*}[!t]
\centering
\small
\setlength{\tabcolsep}{7pt}
\renewcommand{\arraystretch}{1.2}
\caption{Component ablation on N3DV and Meeting Room.
Checkmarks indicate enabled components. The first row
corresponds to IGS (FPS), and the last row to our full
method. Average equally weights the two datasets. LD: Local Deformation}
\label{tab:module_ablation}
\begin{tabular}{cccccccccccc}
\toprule
\multirow{2}{*}{Object}
& \multirow{2}{*}{Flow}
& \multirow{2}{*}{LD}
& \multicolumn{3}{c}{N3DV}
& \multicolumn{3}{c}{Meeting Room}
& \multicolumn{3}{c}{Average} \\
\cmidrule(lr){4-6}
\cmidrule(lr){7-9}
\cmidrule(lr){10-12}
& & &
PSNR $\uparrow$ & DSSIM $\downarrow$ & s/frm $\downarrow$ &
PSNR $\uparrow$ & DSSIM $\downarrow$ & s/frm $\downarrow$ &
PSNR $\uparrow$ & DSSIM $\downarrow$ & s/frm $\downarrow$ \\
\midrule

$\times$ & $\times$ & $\times$
& 32.877 & 0.020 & \best{2.00}
& 30.088 & 0.028 & \best{1.96}
& 31.482 & 0.024 & \best{1.98} \\

$\checkmark$ & $\times$ & $\times$
& 32.882 & 0.020 & 2.20
& 30.116 & 0.028 & 2.14
& 31.499 & 0.024 & 2.17 \\

$\times$ & $\checkmark$ & $\times$
& 32.913 & 0.020 & 2.37
& 30.263 & 0.027 & 2.31
& 31.588 & 0.023 & 2.34 \\

$\times$ & $\times$ & $\checkmark$
& \best{33.059} & 0.020 & 2.28
& 30.109 & 0.028 & 2.24
& 31.584 & 0.024 & 2.26 \\

$\checkmark$ & $\checkmark$ & $\times$
& 32.927 & 0.020 & 2.33
& 30.267 & 0.027 & 2.27
& 31.597 & 0.023 & 2.30 \\

$\times$ & $\checkmark$ & $\checkmark$
& \fair{33.018} & \best{0.019} & 2.29
& \best{30.292} & 0.027 & 2.30
& 31.655 & 0.023 & 2.30 \\

$\checkmark$ & $\times$ & $\checkmark$
& 32.921 & 0.020 & 2.33
& 30.163 & 0.028 & 2.16
& 31.542 & 0.024 & 2.25 \\

\midrule

$\checkmark$ & $\checkmark$ & $\checkmark$
& \good{33.046} & \best{0.019} & 2.30
& \good{30.290} & 0.027 & 2.18
& \best{31.668} & 0.023 & 2.24 \\

\bottomrule
\end{tabular}
\end{table*}











\begin{table}[!t]
\centering
\small
\setlength{\tabcolsep}{5pt}
\renewcommand{\arraystretch}{1.05}
\caption{Ablation of local deformation and anchor
allocation on N3DV and Meeting Room (MR).
Average metrics equally weight the two datasets. LD: Local Deformation}
\label{tab:design_ablation}
\begin{tabular}{@{}lcccc@{}}
\toprule
\multirow{2}{*}{Configuration}
& \multicolumn{2}{c}{PSNR $\uparrow$}
& \multicolumn{2}{c}{Average} \\
\cmidrule(lr){2-3}
\cmidrule(lr){4-5}
& N3DV & MR
& PSNR $\uparrow$
& DSSIM $\downarrow$ \\
\midrule

\multicolumn{5}{@{}l}{\textbf{Local deformation configuration}} \\

No correction
& 32.927 & 30.267
& 31.597 & \best{0.023} \\

Rigid replacement
& 32.988 & 30.204
& 31.596 & 0.024 \\

LD MLP w/o Rigid
& 32.991 & 30.280
& \fair{31.636} & \best{0.023} \\

\midrule
\multicolumn{5}{@{}l}{\textbf{Anchor Reallocation}} \\

FPS
& 32.882
& 30.116 
& 31.499 & 0.024 \\

Object based ($\sqrt{N_o}$)
& \good{33.026} & 30.280
& \good{31.653} & \best{0.023} \\

\midrule
\textbf{Ours}
& \best{33.046} & \best{30.290}
& \best{31.668} & \best{0.023} \\

\bottomrule
\end{tabular}
\end{table}
\vspace{-0.4cm}
\paragraph{Cross-domain Evaluation.}
We evaluate the N3DV-trained checkpoint on Meeting Room,
without further neural-weight training.
Results in \cref{tab:main_results} demonstrate competitive
transfer on Meeting Room.
On Meeting Room, our method achieves 30.290\,dB,
improving over IGS by 0.202\,dB and obtaining lower DSSIM. This shows improved cross-domain transfer of OCLDA-GS.

\paragraph{Reconstruction Efficiency.}
As discussed in \cref{sec:prelim}, we do
Gaussian optimization only at keyframes and use neural
prediction for intervening frames.
On N3DV and Meeting Room, reconstruction requires
2.30 and 2.18\,s/frm, respectively, compared with
2.00 and 1.96\,s/frm for IGS.
The added processing cost is approximately 0.30 and
0.22\,s/frm, accompanying the quality improvements on
these datasets.
Across the completed comparisons in \cref{tab:main_results},
our method has the second-lowest reported reconstruction
time on every dataset.
Our method achieves an average PSNR of 29.045\,dB
with 594 total refinements, outperforming IGS at the
same refinement count and achieving the highest average
PSNR among online methods using the fewest refinements.
QUEEN, the only online method with higher average PSNR
(29.486\,dB), uses 2,994 refinements, approximately
5.04$\times$ as ours.

\subsection{Qualitative Results}

\cref{fig:qualitative} compares our method with online baselines
on selected examples from the five datasets. Our reconstructions
show closer alignment with the ground truth around moving objects,
with reduced ghosting and boundary artifacts, complementing the
aggregate comparisons in \cref{tab:main_results}.

\cref{fig:motion_propagation} illustrates anchor selection,
group propagation, and local deformation correction on a Meeting Room
sequence. Flow over the completed interval reallocates a fixed
anchor budget at the current keyframe. Object group-constrained
propagation preserves the recovered grouping, while the learned
local deformation adjusts the propagated deformation.
The final panel shows the reconstruction after standard
target-keyframe refinement.
\begin{figure}[t]
    \centering
    \includegraphics[width=\linewidth]{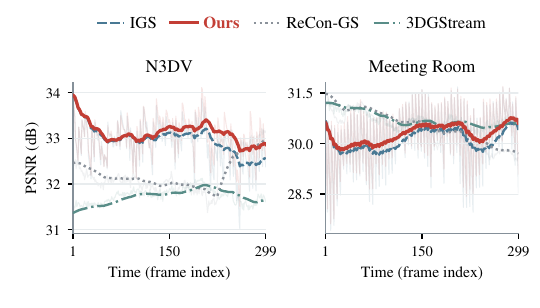}
\caption{\textbf{Temporal reconstruction quality with selected online baselines.}
Faint traces show raw values; solid curves show a 45-frame EMA.
Full comparisons appear in \cref{tab:main_results}.}
    \label{fig:temporal_backbone}
\end{figure}

\begin{figure}[t]
\centering
\includegraphics[width=\linewidth]{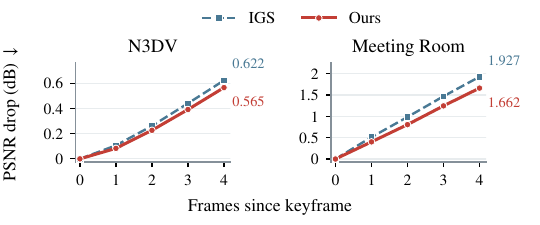}
\caption{Mean PSNR drop relative to keyframes on N3DV
and Meeting Room; lower is better.}
\label{fig:temporal_drift}
\end{figure}

\begin{figure}[t]
\centering
\includegraphics[width=\linewidth]{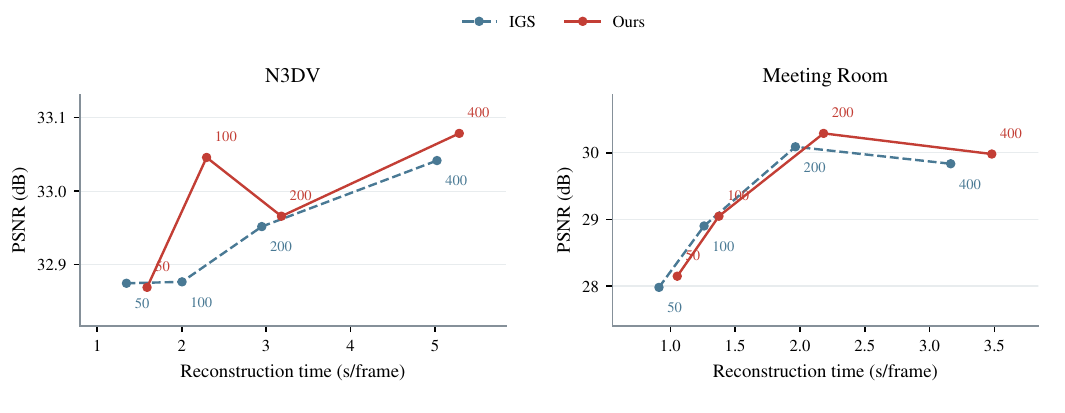}
\caption{\textbf{Refinement budget and reconstruction quality.} Numbers denote keyframe refinement steps.}
\label{fig:refinement_budget}
\end{figure}

\subsection{Ablation Study}

\paragraph{Component contributions.}
\cref{tab:module_ablation} compares flow-guided anchor selection,
object-constrained propagation, and learned local deformation correction.
Object restricts motion propagation within object groups, Flow prioritizes
moving objects during anchor allocation, and LD learns per-anchor corrections.
The complete method improves PSNR over IGS by 0.169\,dB on N3DV
and 0.202\,dB on Meeting Room.
Flow-containing variants consistently improve Meeting Room
reconstruction, while local deformation correction provides the highest
N3DV PSNR when combined with FPS.
Although component gains are not uniformly additive, the complete
method remains within 0.013\,dB and 0.002\,dB of the highest-scoring
configurations on N3DV and Meeting Room, respectively.

\AD{\cref{tab:design_ablation} compares Local Deformation (LD) and anchor
reallocation. For LD configuration, No correction disables LD, while
Rigid replacement uses estimated rigid motion only.
LD MLP w/o Rigid learns per-anchor corrections
without conditioning them on the estimated rigid rotations.
For anchor reallocation, FPS selects anchors for spatial coverage,
while Object based ($\sqrt{N_o}$) allocates anchors
by object size without motion guidance.
Ours combines LD conditioned on rigid motion
with motion-guided anchor reallocation, achieving the
highest average PSNR of 31.668\,dB.}


\paragraph{Cross-domain temporal robustness.}
Meeting Room is excluded from local deformation-head training, so its
improvement over IGS demonstrates transfer beyond the training
datasets without retraining the head.
Beyond average quality, \cref{fig:temporal_drift} shows smaller
PSNR degradation between keyframes.
The mean drop in PSNR from frames 0 to 4 from keyframe decreases from
0.622 to 0.565\,dB on N3DV and from 1.927 to 1.662\,dB
on Meeting Room, corresponding to reductions of 9.2\% and 13.8\%.
Together with the full-sequence trajectories in
\cref{fig:temporal_backbone}, these results support improved
quality preservation between refinement updates, including
on the unseen dataset.

\paragraph{Number of anchors.}
On cross-domain dataset Meeting Room evaluation, OCLDA-GS
outperforms IGS by 0.113 dB using eight times fewer
motion anchors. \cref{tab:anchor_efficiency} shows consistent PSNR gains
on N3DV, Meeting Room, and Google Immersive across
all anchor budgets. \AD{This further demonstrates the advantage of anchor re-allocation under tighter budgets.}

\begin{table}[!t]
\centering
\footnotesize
\setlength{\tabcolsep}{1.75pt}
\renewcommand{\arraystretch}{1.15}
\caption{Motion-anchor budget ablation across the datasets. PSNR ($\uparrow$) averages evaluated
targets per scene, followed by an equal-scene average
within each dataset. Shading indicates the highest PSNR
per dataset.}
\label{tab:anchor_efficiency}
\begin{tabular}{@{}lcccccc@{}}
\toprule
\multirow{2}{*}{Dataset}
& \multicolumn{2}{c}{1,024}
& \multicolumn{2}{c}{4,096}
& \multicolumn{2}{c}{8,192} \\
\cmidrule(lr){2-3}
\cmidrule(lr){4-5}
\cmidrule(lr){6-7}
& IGS & Ours & IGS & Ours & IGS & Ours \\
\midrule
N3DV
& 32.741 & 32.822
& 32.851 & 32.962
& 32.877 & \best{33.046} \\
Meeting Room
& 29.860 & 30.201
& 30.057 & 30.263
& 30.088 & \best{30.290} \\
Google Immersive
& 23.637 & 23.766
& 23.593 & 23.598
& 23.599 & \best{23.798} \\
\bottomrule
\end{tabular}
\end{table}

\paragraph{Refinement efficiency.}
\cref{fig:refinement_budget} examines the PSNR--runtime tradeoff
across four keyframe refinement budgets.
On N3DV, OCLDA-GS with 100 refinement steps achieves
32.995\,dB at 2.13\,s/frame, compared with
32.952\,dB at 2.95\,s/frame for IGS with 200 steps.
This yields slightly higher PSNR with 27.8\% less reconstruction
time.
On Meeting Room, Ours with 200 steps reaches 29.777\,dB,
within 0.057\,dB of IGS with 400 steps, while taking
33.5\% less time.
These results indicate that the proposed propagation
can reduce the refinement budget needed to obtain comparable PSNR.

\section{Limitations}
OCLDA-GS depends on the quality of object groups.
Inaccurate
grouping can limit object-aware motion propagation. More robust object grouping
could alleviate this limitation. Similarly, the rigid-motion mechanism assumes that each object group results in a useful dominant rigid transformation, which may be inaccurate for articulated objects. Finally, our learned local deformation correction updates Gaussian positions only but does not correct errors in orientation, scale, and appearance.    
\section{Conclusions}
We presented OCLDA-GS, an object-centric Gaussian streaming
framework that combines flow-guided anchor selection,
object-constrained motion propagation, and learned
local deformation correction conditioned on object-wise rigid
motion. Under a fixed anchor budget, these components
improve motion modeling without per-frame Gaussian
refinement. Experiments across three dynamic multi-view
datasets demonstrate competitive reconstruction quality
and efficiency, with improvements over IGS on N3DV
and Meeting Room and reduced quality degradation
between keyframes. These results highlight the importance
of object-aware motion modeling for efficient Gaussian
streaming.    
\clearpage
\section*{LLM Usage Statement}
OpenAI ChatGPT was used to obtain reviewer-style feedback on the manuscript or major portions of it. It assisted with editing and refining author-produced text, including related work and citation presentation, and with refining author-produced material concerning experimental design, execution, analysis of results, and figures or diagrams. For code and implementation, ChatGPT generated work that the authors reviewed, edited, or approved. No LLMs or generative AI tools were used for synthetic data, dataset creation, annotation, or developing research ideas, hypotheses, claims, or conclusions. The authors take full responsibility for all submitted content, regardless of any LLM or AI-tool use.

{
    \small
    \bibliographystyle{ieeenat_fullname}
    \bibliography{main}
}


\end{document}